\documentclass[pdflatex,sn-mathphys-num]{sn-jnl}

\usepackage{graphicx}%
\usepackage{multirow}%
\usepackage{amsmath,amssymb,amsfonts}%
\usepackage{amsthm}%
\usepackage{mathrsfs}%
\usepackage[title]{appendix}%
\usepackage{xcolor}%
\usepackage{textcomp}%
\usepackage{manyfoot}%
\usepackage{booktabs}%
\usepackage{algorithm}%
\usepackage{algorithmicx}%
\usepackage{algpseudocode}%
\usepackage{listings}%
\usepackage{graphicx}
\usepackage[numbers]{natbib}
\usepackage{multirow}

\usepackage{subcaption}
\usepackage{amsmath}

\usepackage{float}
\usepackage{amsthm}
\usepackage{placeins}
\usepackage{algpseudocode}
\usepackage{algorithm}

\usepackage{array}

\theoremstyle{thmstyleone}%
\theoremstyle{thmstyletwo}%

\theoremstyle{thmstylethree}%

\begin{document}
\title[Article Title]{Transparent AI for Mathematics: Transformer-Based Large Language Models for Mathematical Entity Relationship Extraction with XAI}


 \author*[1]{\fnm{Tanjim Taharat} \sur{Aurpa}}\email{aurpa0001@uftb.ac.bd}

 \affil*[1]{\orgdiv{Department of Data Science and Engineering}, \orgname{University of Frontier Technology, Bangladesh}, 
 }


\abstract{Mathematical text understanding is a challenging task due to the presence of specialized entities and complex relationships between them. This study formulates mathematical problem interpretation as a Mathematical Entity Relation Extraction (MERE) task, where operands are treated as entities and operators as their relationships. Transformer-based models are applied to automatically extract these relations from mathematical text, with Bidirectional Encoder Representations from Transformers (BERT) achieving the best performance, reaching an accuracy of 99.39\%. To enhance transparency and trust in the model’s predictions, Explainable Artificial Intelligence (XAI) is incorporated using Shapley Additive Explanations (SHAP). The explainability analysis reveals how specific textual and mathematical features influence relation prediction, providing insights into feature importance and model behavior. By combining transformer-based learning, a task-specific dataset, and explainable modeling, this work offers an effective and interpretable framework for MERE, supporting future applications in automated problem solving, knowledge graph construction, and intelligent educational systems.}




\maketitle

\section{Introduction}
Academicians nowadays use different modern technology to conduct their academic activities. Basic educational subjects like Mathematics, Science, and language-related subjects require significant attention in order to develop such a system, and it is essential that the systems are able to process specialized texts such as mathematical problems containing different mathematical entities, their relationships, and other words. MERE focuses on identifying mathematical entities from text and determining relationships between them. A math problem can be considered as an entity-entity relationship problem, such as the operands being the entities and the operator being the relation between them.
This task bridges the gap between natural language and mathematical notation, enabling systems to understand and extract meaningful relationships between mathematical concepts in documents like research papers, textbooks, or theorems. The potential of MERE in advancing mathematical knowledge management, semantic search, and educational tools is promising, offering a bright future for its applications. 

In the realm of modern NLP research, transformer-based models like BERT, ELECTRA, XLNet, etc., have garnered significant attention for their performance. BERT, in particular, has been a game-changer. BERT, short for BERT, is a pre-trained transformer that can understand multiple languages and can be applied to various downstream tasks. Developed by Google, BERT captures context from both directions of a sentence, a feature that makes it exceptionally effective for tasks such as question answering \cite{aurpa2022reading, aurpa2023uddipok}, text classification \cite{aurpa2021progressive, kulkarni2021experimental, krishnan2021cross}, entity recognition \cite{aurpa2024ensemble}, etc. Its use of the Transformer architecture and its ability to fine-tune downstream tasks have revolutionized NLP, setting new standards for performance.

XAI for BERT involves techniques like SHAP that help interpret the model's predictions by highlighting the influence of each input token on the final output. These methods make it easier to understand which parts of a sentence contribute the most to BERT's decisions, offering transparency to its "black-box" nature. The use of SHAP can be seen in text classification \cite{kokalj2021bert, tao2023making}, multimodal prediction \cite{van2024feature, el2021multilayer}, image processing \cite{ukwuoma2025enhancing} etc.
For MERE, XAI can be used to pinpoint which words or mathematical symbols are most critical in identifying entities and their relationships, thus improving interpretability and trust in automated extraction systems from complex mathematical texts. 

In this work, a mathematical problem has been treated as an entity relationship problem, with the entities as the operands and the relationship among them as the operator. Given the current technological landscape, this research aims to use BERT to extract entity-entity relations from mathematical texts and then interpret and explain our model using SHAP, a tool that provides insights into the model's decision-making process. The main objectives of this paper are as follows:
\begin{itemize}
    \item Propose an approach that considers math problems as entity relationship problems.
    \item Develop a BERT-based technique to extract relationships between mathematical entities.
    \item Utilizing the Explainability method named SHAP to understand the contribution of the feature in the prediction.
\end{itemize}

\section{Related Work}
BERT has been widely adopted by many researchers due to its exceptional performance in addressing various challenges. In \cite{li2019exploiting}, the study highlights the use of BERT for contextual embedding modeling. By incorporating a simple linear layer for classification, they modified BERT's architecture, which led to performance improvements. In another study, \cite{liu2019bb}, BERT is employed in a context-aware application, showing an average improvement of 1.65\% over existing CNN and BiLSTM-based methods. Similarly, \cite{utka2020pretraining} conducted research on a Twitter-based dataset, pre-training BERT to enhance Latvian sentiment analysis on tweets. Additionally, BERT is also prominent in fake news detection. In \cite{rai2022fake}, the authors proposed a hybrid approach that combines BERT with LSTM by connecting BERT's output to an LSTM layer. This model increased accuracy by 2.50\% on the PolitiFact dataset and by 1.10\% on the GossipCop dataset. Authors in \cite{aurpa2024ensemble} ensembled two BERT models for the recognition of mathematical entities and remarkably improved accuracy by 1.78\%. 

Across recent studies, SHAP is consistently positioned as a core interpretability mechanism that bridges high-performing transformer models with human-understandable insights, regardless of domain.
In traffic safety analysis comparing the USA and Jordan, SHAP enables the identification of culturally, infrastructurally, and legally grounded risk factors embedded in crash narratives, ensuring that BERT-based predictions are not treated as black-box outputs but as explainable, policy-relevant findings \cite{jaradat2025cross}. In multi-label emotion classification, SHAP (alongside LIME) serves as a benchmark for explanation quality, revealing that while statistical explanation methods highlight influential tokens, generative AI models provide richer, more semantically aligned reasoning, and insight quantitatively supported through BERTScore comparisons \cite{alyoubi2025interpretable}. In VANET intrusion detection, SHAP shifts from text-level interpretation to feature-level transparency, clarifying how specific safety-message attributes contribute to identifying false positional information in highly dynamic vehicular environments, thereby increasing trust in safety-critical systems \cite{khan2025novel}. Finally, in aspect-based sentiment analysis, SHAP is explored at different stages of the representation pipeline, demonstrating that the point of SHAP application (before versus after contextual embedding) significantly affects the granularity and faithfulness of explanations \cite{yeung2024explaining}. 
Authors in \cite{siddiqui2024towards} SHAP for the classification of emotion using different transformer-based architectures and explain the significant features for the prediction. Another work \cite{zhao2020shap} utilized SHAP to visualize the explainability for CNN-based text classification. They used the Amazon Electronic Review dataset and scored the crucial features for classification. Rabbi et al. used SHAP for multi-label text classification with COVID-19-related text, and here, the highest-performing classifiers were Random Forest and BERT \cite{rabby2023multi}. Shap is also utilized to recognize fake and authentic news and question classification in \cite{tao2023making}, where the author worked based on binary, multi-class, and multi-label classification. Another use \cite{dewi2022shapley} can be seen for the classification of the internet movie dataset, where authors also consider the sentiment analysis for showing the explainability with SHAP. Collectively, these works establish SHAP as a flexible, domain-agnostic XAI framework that enhances transparency, validates model behavior, and supports the real-world deployment of complex AI systems across transportation safety, emotion analysis, vehicular cybersecurity, and sentiment analysis.

Several works have been conducted on mathematical entity-entity relation extraction. In \cite{jie2022learning}, authors proposed a deductive reasoning process, ROBERTA-DEDUCT REASONER, which thinks of math problems as  Entity entity Relation Extraction. This approach provided about 92\% accuracy and presented deduced reasoning for the analysis of complex mathematical relations. Another work \cite{lin2021hms} proposed a hierarchical mathe solver(HMS) where the math statement is first departed into several divisions and semantic information from each part to solve the entire statement as the encoder. The authors used two different datasets named Math23K and MAWPS, and the proposed method outperformed them with 0.761 and 0.803, respectively. Wu et al. \cite{wu2020knowledge} proposed a knowledge-aware sequence tree (KA-S2T) using the Math23K dataset to achieve a better performance. This KA-S2T outperformed with 76.3\% accuracy. Pretraining Language Modeling is utilized in \cite{liang2022mwp} where the author proposed MWP-BERT and MWP-RoBERTa for solving Math Word Problems. For the Math23k and Math23k datasets, MWP-BERT showed the best performance with 84.7 and \% 82.4\% accuracy scores, and MWP-RoBERTa performed best in the MathQA dataset with 76.6\% accuracy. Authors in \cite{liu2020mathematical} proposed a generative model to create Math Word problems that work based on gated graph neural networks (GGNNs) and variational autoencoder (VAE). They used the large-scaled MWP dataset with outperformed standard deviation. Embedding-based approaches have been shown to be effective for similarity-based retrieval of mathematical expressions, where unimodal and multimodal deep learning models operating on graph and image representations learn high-dimensional embeddings optimized through metric learning and hard-sample mining, enabling efficient and accurate retrieval compared to traditional scoring functions \cite{ahmed2021equation}. Complementing retrieval tasks, explainable document subject classification has gained attention in STEM domains, where the integration of textual and mathematical entity linking provides interpretable insights into classification decisions, addressing the limitations of black-box models and demonstrating that mathematical entities play a crucial role in enhancing explainability for complex technical documents \cite{scharpf2021towards}. In parallel, research on math word problem solving has shifted toward explainable reasoning frameworks, framing the task as a relation extraction problem that explicitly models deductive reasoning between quantities through stepwise primitive operations, thereby improving both predictive accuracy and transparency over sequence-based generation methods \cite{jie2022learning}. Collectively, these studies highlight a growing emphasis on combining deep representation learning with explainability to support retrieval, classification, and reasoning in mathematical and STEM-centric applications.

Several works have been conducted on math entities and their relationship recognition, but large language models are not significantly utilized in this field. Moreover, XAI techniques are not focused on such problems to verify the transparency of the model's performance. Therefore, the importance of the research questions raised and addressed in this research cannot be overstated. The research questions are:
\begin{itemize}
    \item[1] How effectively can large language models such as transformer-based architectures, such as BERT, be adapted for relation extraction in mathematical texts?
    \item[2] What is the impact of XAI techniques, such as SHAP, on improving the transparency and interpretability of automated relation extraction models in mathematical contexts?
    \item[3] How does the proposed relation extraction model perform in terms of accuracy and other evaluation scores?
    \item[4] How can deep learning techniques utilized for math entity relationship extraction be benefited in modern automated systems?
\end{itemize}
This research addresses the previously unresolved challenge of MERE by introducing a novel and principled framework grounded in modern deep learning techniques. Unlike existing approaches, which largely rely on traditional or heuristic-based methods, the proposed work uniquely leverages state-of-the-art transformer architectures to model complex semantic and structural relationships between mathematical entities in text. The study further contributes a dedicated dataset tailored to this task, filling a critical resource gap and enabling systematic evaluation and reproducibility. In addition to advancing predictive performance, the research incorporates XAI techniques to interpret model decisions, providing transparent insights into how textual and mathematical cues influence relation inference. Collectively, these contributions establish the originality and novelty of the work by unifying transformer-based representation learning, dataset creation, and explainability within a single, coherent framework for mathematical entity relation analysis.

\section{Proposed Methodology}

In this section, the preliminaries of our methodology and the proposed approach used here is discussed. The major part of the methodology of this research is explained below:

\subsection{Data Collection and Preprocessing} 
\label{sec:dp}
The dataset of this research is developed utilizing two different datasets mentioned as follows:
\begin{itemize}
    \item Bangla\_MER\cite{aurpa2024bangla_mer}: This dataset contains both Bangla and English texts for Mathematical Entity Recognition. Here, only the English dataset is considered with columns Text, Entity Name, and Entity Types.
    \item Somikoron\cite{aurpa2024shomikoron}: While creating this dataset, the authors at first collected the mathematical statements from the Bangla\_MER dataset, then determined the relevant text for each statement.
\end{itemize}
\begin{figure}
    \centering
    \includegraphics[width=1.2\linewidth]{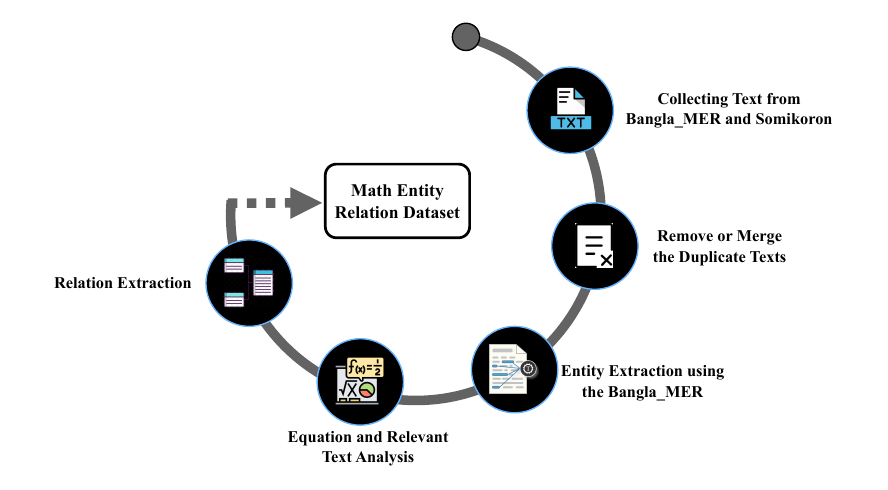}
    \caption{Step-by-step pipeline for constructing the Mathematical Entity–Entity Relation Dataset, illustrating text collection from Bangla\_MER and Somikoron datasets, duplicate handling, entity extraction, equation analysis, and manual relation labeling.}
    \label{data_creation}
\end{figure}

The dataset creation process begins by collecting texts from the dataset mentioned. Then duplicate observations are removed or merged. Then the entities are extracted from the mathematical text. While manually extracting entities, it was ensured that the chosen mathematical phrases actually function as operands in the text. In this dataset, we have worked with number phrases instead of words. For example, 'five thousand and forty' has been treated as one mathematical entity. This step has been conducted using the Bangla\_MER dataset. Next, the equations of the corresponding text were analyzed from the 'Somikoron' Dataset and 
The relations were manually checked, and the data set was labeled. Figure \ref{data_creation} depicts the entire process of collecting the dataset.

Some observations from the provisioned dataset are given in Table \ref{tab:dataset}.

\begin{table*}[]
\centering
\caption{The observations of the developed dataset by combining Bangla\_MER and Somikoron Datasets.}
\label{tab:dataset}
\begin{tabular}{|l|c|c|c|}
\hline
\multicolumn{1}{|c|}{\textbf{Text}} & \textbf{Entity 1} & \textbf{Entity 2} & \textbf{Relationship} \\ \hline
\begin{tabular}[c]{@{}l@{}}Eighteen players of the Sri Lankan\\national cricket team have come to \\play in Bangladesh. Bangladesh team \\also has eighteen players. There is \\total thirty six players in two teams.\end{tabular} & Eighteen & Eighteen & Addition \\ \hline
\begin{tabular}[c]{@{}l@{}}Subtracting fifty from ten is equals \\forty.\end{tabular} & Fifty & Ten & Subtraction \\ \hline 
\begin{tabular}[c]{@{}l@{}}Two thousand multiplied by five is \\ten thousand. \\ \end{tabular} & Two Thousand & Two & Multiplication \\ \hline
The square root of four is two  . & Four & Two & Square Root \\ \hline
\begin{tabular}[c]{@{}l@{}}The factorial value of seven is \\five thousand and forty.\end{tabular} & Seven & Five Thousand Forty & Factorial \\ \hline
\begin{tabular}[c]{@{}l@{}}Man bought twelve mangoes and \\divided them equally among three \\children. Each child got four of \\mangoes.\end{tabular} & Twelve & Three & Division \\ \hline
\end{tabular}
\end{table*}

\begin{figure}
    \centering
    \includegraphics[width=0.8\linewidth]{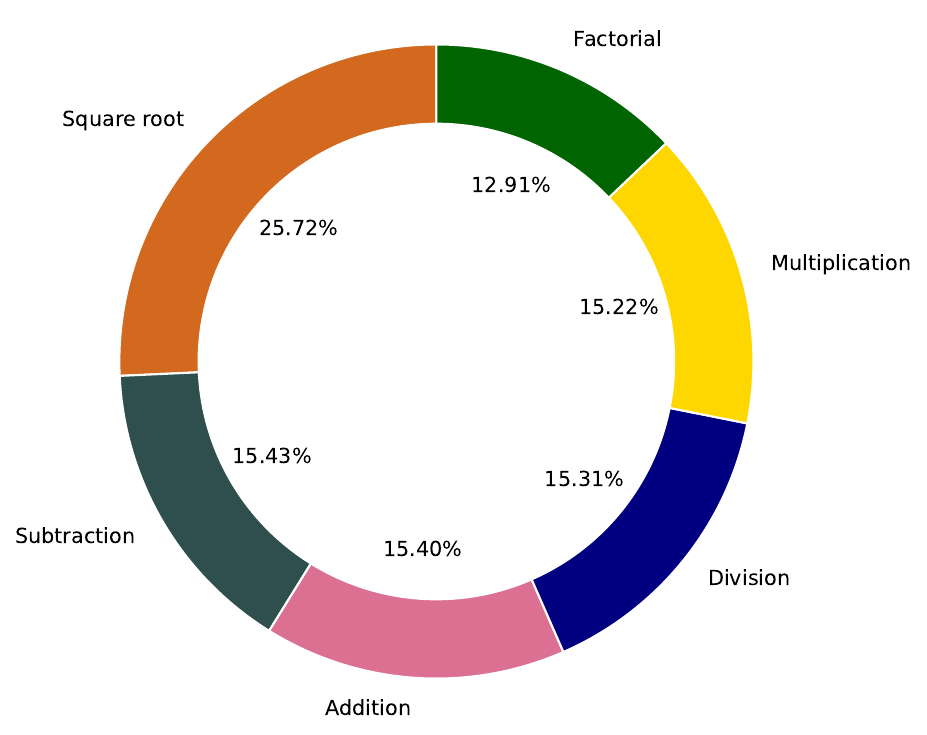}
    \caption{Donut chart showing the percentage distribution of mathematical relation classes in the constructed dataset, highlighting the dominance of subtle Square Root operations and the balanced representation of Addition, Subtraction, Multiplication, and Division.}
    \label{classdis}
\end{figure}

The dataset consists of a total of 3284 unique mathematical statements. In Figure \ref{classdis}, the donut chart shows the percentage distribution of different mathematical relations, where the Square root is the most frequently used relation at about 25.72\%, while Factorial is the least used at around 12.91\%. The remaining relations—Addition, Subtraction, Division, and Multiplication—each contribute nearly equal shares, all close to 15\%, indicating a fairly balanced usage among all the relations. Overall, the chart suggests that most operations are used at similar rates, with square root standing out as the dominant one.

Raw data often includes unnecessary letters and words, which may make categorization difficult. Several preprocessing procedures are performed before feeding the data into the classifier to guarantee accurate categorization \cite{ahmed2020online}. The following actions are necessary for getting the most significant results:

\begin{itemize} 
\item The raw dataset includes various non-letter characters (such as \$, \%, \#, \*, and -), which may negatively influence accuracy. Thus, these characters are removed from the dataset. 

\item Additionally, the data contains many stop words that do not contribute to the prediction tasks and can lower precision. Removing these stop words has enhanced the accuracy of our model. We remove the stopwords collected from NLRK\footnote{https://www.nltk.org/}.

\item Lemmatization reduces words to their base or dictionary form (lemma) while stemming cuts off word endings to return a root form, often producing non-standard words. Lemmatization and stemming are used to reduce token size and the model's complexity.
\end{itemize}

\begin{figure}
\centering
    \includegraphics[width=.9\linewidth]{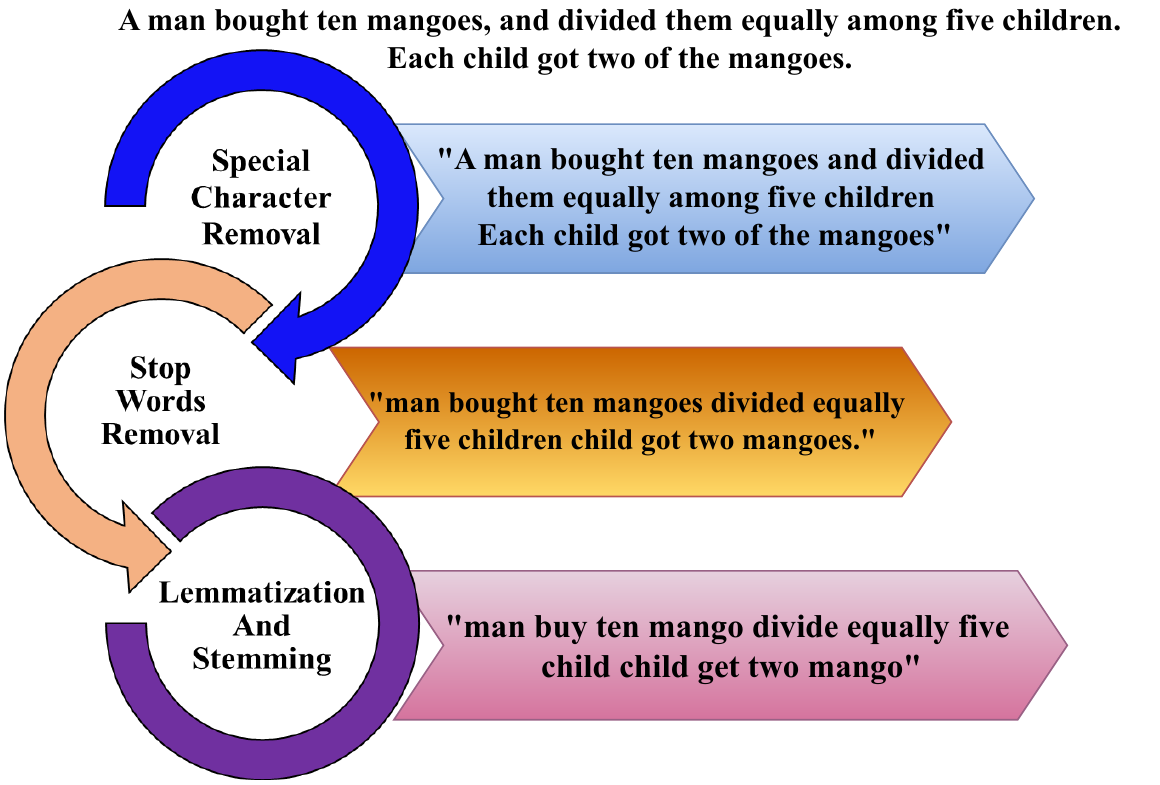}
    \caption{Illustration of the data preprocessing workflow, demonstrating raw text normalization through character cleaning, stop-word removal, lemmatization, and stemming to generate model-ready mathematical text.}
    \label{fig:data preprocess}
\end{figure}

Figure \ref{fig:data preprocess} depicted this research work's entire data preprocessing method.

\subsection{Model Building, Training and Evaluation}
After completing data collection and preprocessing, the model has been built and trained. Here, model building, training, and evaluation are briefly described.

\subsubsection{Proposed BERT Model}
\begin{figure}
    \centering
    \includegraphics[width=0.8\linewidth]{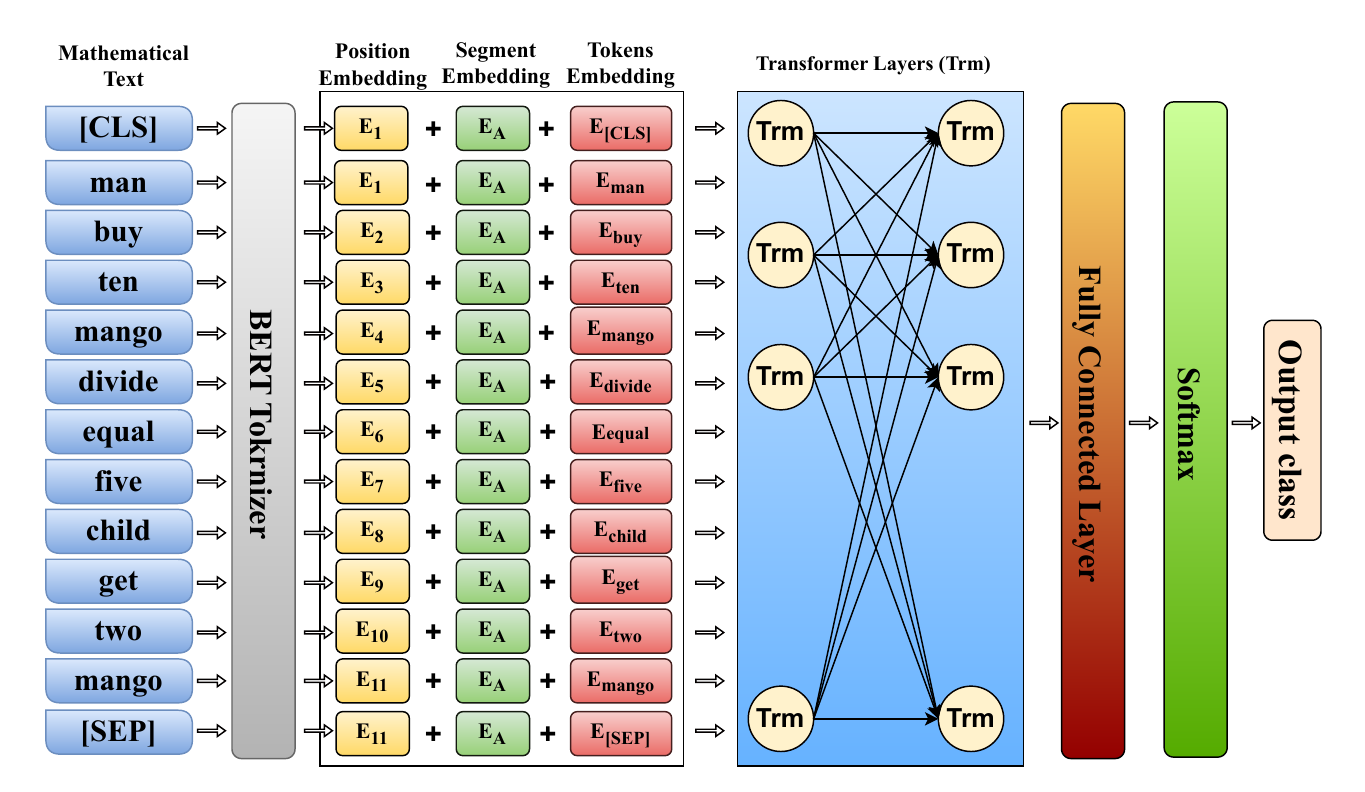}
    \caption{Proposed BERT Model}
    \label{fig:bert model}
\end{figure}

In this work, BERT \cite{devlin2018bert} have been used. This is a transformer-based architecture that uses the attention mechanism. BERT has two main parts discussed below:

\begin{itemize}
\item \textbf{Pre-training BERT: } BERT is initially trained on two unsupervised learning tasks: Masked Language Modeling (MLM) and Next Sentence Prediction (NSP). In MLM, a subset of random tokens is masked, and the model predicts them, resulting in a pre-trained bidirectional representation. NSP is designed to predict whether a particular sentence follows another in a pair, which helps the model understand sentence relationships when multiple sentences are provided. BERT’s pre-training utilized text passages from English Wikipedia (excluding lists, headings, and tables) and the BooksCorpus, which contains 800 million words \cite{zhu2015aligning}. The 'bert-base-uncased' (HuggingFace Transformers) checkpoint has been used as the pretrained model here.

\item \textbf{Fine-tuning BERT: } BERT is widely used for various downstream tasks involving single sentences or paired sentences as inputs. It starts with pre-trained parameters, which are then fine-tuned for specific tasks using labeled data.
\end{itemize}
Figure \ref{fig:bert model} depicts the proposed Bert model. At first, the preprocessed mathematical text is passed through the BERT tokenizer and generates attention mask(positional encoding), segment embedding, and token embedding. Then, following the BERT layer with 24 transformers, these sequences are sent to a fully connected layer, and then finally, with prediction, the output relation class is extracted. The 80\% data has been used to train the model, and the remaining data is used for testing purposes. 
While splitting the dataset into a test and a train set, data were randomly chosen for each set. There were no duplicate statements in the entire dataset.

\subsubsection{Hyperparameter Tuning}
Altering the values of the model's different hyperparameters can affect the performance by changing its capacity, convergence rate, and how the model responds to data patterns. Careful tuning and experimentation, which involves testing various combinations of hyperparameters and observing the results, are crucial to optimizing model accuracy for specific tasks. Overall, changing hyperparameters influences a model's accuracy by impacting its capacity, generalization, optimization dynamics, and sensitivity to data features. Therefore, the BERT model is fine-tuned using multiple combinations of learning rate, verbose, batch size, epoch, and max length, and the findings of this research are validated through rigorous experimentation to achieve optimal performance. For the following values in Table \ref{tab:hyppara}, the model provided the best performance.

\begin{table}
\caption{This table lists the model parameters and their corresponding values.}
\label{tab:hyppara}
\begin{tabular}{|l|l|}
\hline
\multicolumn{1}{|c|}{\textbf{Hyperparameters}} & \multicolumn{1}{c|}{\textbf{BERT}} \\ \hline
learning rate (AdamW) & 2e-04 \\ \hline
max\_len & 50  \\ \hline
Batch\_size & 12 \\ \hline
verbose & 1 \\ \hline
epoch & 40 \\ \hline
\end{tabular}
\end{table}

The model has been trained with the different values of hyperparameters, and the values in Table \ref{tab:hyppara} showed the highest performance. Therefore, these values are selected for this research. Since the text lengths in the dataset are no more than 50, max\_len has been set to 50 to reduce the model's complexity.

\subsubsection{Model Evaluation}
For the evaluation of our model, this research utilized a variety of evaluation metrics, such as accuracy and micro and macro average F1 score. The calculation of the confusion matrix played a pivotal role in determining these metrics. The following Equations \ref{acc}, \ref {microFI} and \ref {MacroFI} were used to compute the accuracy and micro and macro average F1 score.

\begin{equation}
    \text{Accuracy} = \frac{TP + TN}{TP + TN + FP + FN}
    \label{acc}
\end{equation}

\begin{equation}
    \text{Micro F1 Score} = \frac{\sum TP}{\sum TP + \frac{1}{2} \left(\sum FN + \sum FP\right)}
    \label{microFI}
\end{equation}

\begin{equation}
    \text{Macro FI Score} = \frac{\sum_{i=1}^{\text{ No of classes}} \text{F1 Score}_i}{\text{ No of classes}}
    \label{MacroFI}
\end{equation}

Here, TP, TN, FP, and FN indicate True Positive, True Negative, False Positive, and False Negative, respectively. Some other equations have also been used in this paper, and they are mentioned below. Equation \ref{prec}, \ref{recall}, \ref{specificity}, \ref{err} is used to determine Precision, Recall, Specificity, and Error Rate.

\begin{equation}
    \text{Precision} = \frac{\text{TP}}{\text{TP} + \text{FP}} \times 100\%
    \label{prec}
\end{equation}

\begin{equation}
    \text{Sensitivity/Recall} = \frac{\text{TP}}{\text{TP} + \text{FN}} \times 100\%
    \label{recall}
\end{equation}

\begin{equation}
    \text{Specificity} = \frac{TN}{TN + FP}
    \label{specificity}
\end{equation} 

\begin{equation}
    \text{Error Rate} = \frac{FP + FN}{TP + TN + FP + FN}
    \label{err}
\end{equation}
These measures are computed based on the validation dataset, which is 20\% of the main dataset. Due to the lack of existing data, this model can not be evaluated on paraphrased problems or problems with wrong mathematical statements, but the future research plan includes the inclusion of such an evaluation process.

\subsection{Model Explainability}
SHAP is a unique and valuable tool in our field, as it has been utilized to explain and understand our model's important features. This popular framework for interpreting the output of machine learning models assigns each feature an importance value for a given prediction based on Shapley values from cooperative game theory. These values provide a fair distribution of payoff among players (or features) in a coalition. In our context, SHAP's unique contribution lies in its ability to elucidate how different features of mathematical text contribute to the identification and extraction of entities.
This paper's approach integrates SHAP as a key component to interpret the outputs of the entity relationship extraction model tailored for mathematical text. By leveraging SHAP values, the model's insights about the decision-making process, particularly:
\begin{itemize}
\item Feature Importance: SHAP highlights which features (e.g., words) are most influential in the accurate extraction of entities.
\item Model Transparency: The utilization of SHAP in this work enhances the transparency of our model. By allowing us to understand why certain predictions are made, SHAP builds trust in the model's reliability, making it a valuable tool for practitioners in the field.
\item Error Analysis: SHAP aids in diagnosing errors with pinpoint precision, identifying features that lead to incorrect predictions and facilitating targeted model improvements.
\end{itemize}
\begin{figure}
\centering
    \includegraphics[width=0.5\linewidth]{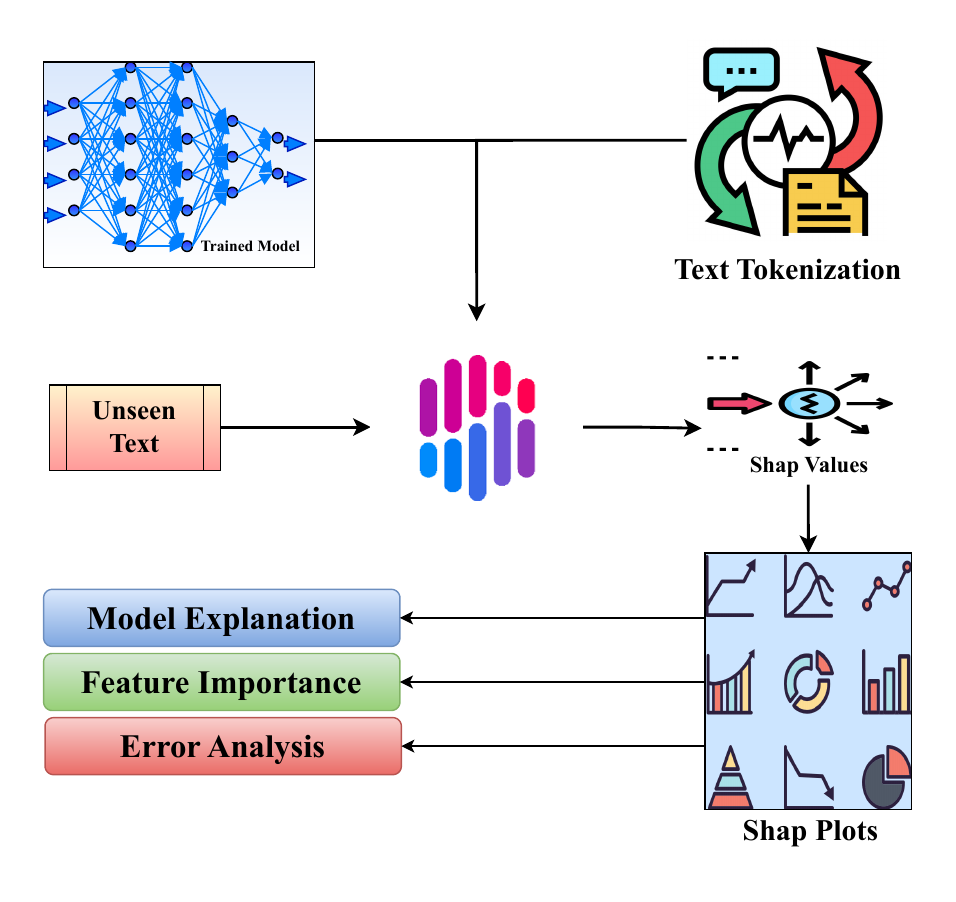}
    \caption{The working process of SHAP}
    \label{fig:shapfig}
\end{figure}
Figure \ref{fig:shapfig} illustrates the working principle of the SHAP algorithm. Here we have used shap.Explainer(pred). This call automatically selects a model-agnostic SHAP explainer because the input is a Hugging Face transformers pipeline, which is treated as a black-box prediction function. Instead of accessing internal model gradients or attention weights, SHAP estimates feature contributions by perturbing input tokens and observing changes in the output probabilities returned by the text-classification pipeline. This approach follows the principles of Shapley values from cooperative game theory, assigning each token a contribution score toward the final prediction. As a result, the explanations are faithful, model-independent, and applicable across transformer architectures, though computationally more expensive than gradient-based SHAP variants.

\subsection{Proposed Approach}
\begin{figure*}
    \centering
    \includegraphics[width=0.7\linewidth]{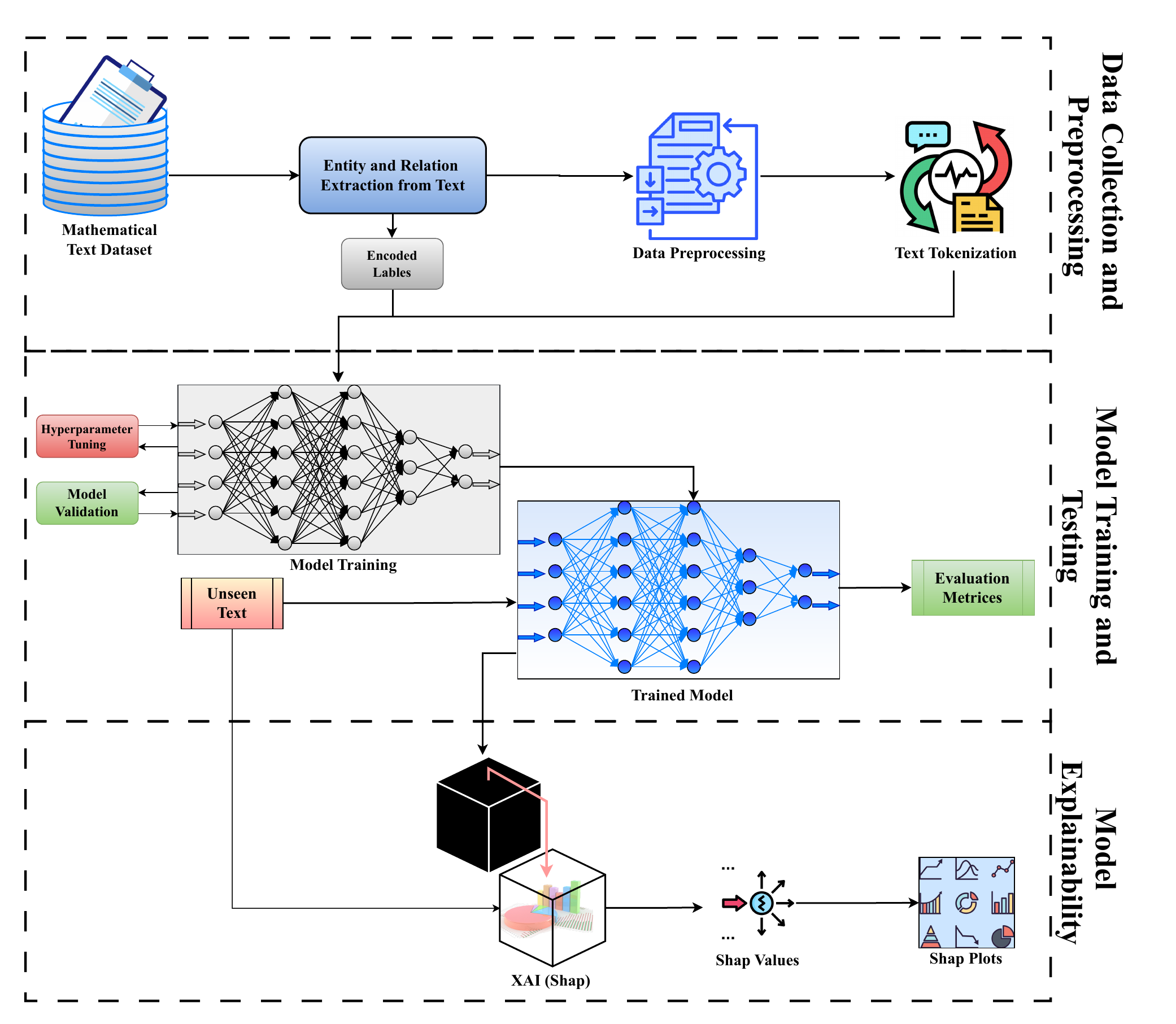}
    \caption{Overall workflow of the proposed approach, integrating data collection and preprocessing, transformer-based model training and evaluation, and SHAP-based explainability for transparent mathematical relation extraction.}
    \label{fig:sys_arc}
\end{figure*}

Figure \ref{fig:sys_arc} delineated the workflow diagram of the approach that has been proposed here. The whole workflow can be divided into three distinct parts. In the first division, a thorough and step-by-step data collection and preprocessing process is outlined, where the entities and relations between entities are extracted, the text is preprocessed and tokenized, and labels are encoded. Next, the second division is about model building and training, including essential steps like hyperparameter tuning and model evaluation. At this stage, a trained model is obtained and able to predict unseen data. Finally, the Explainability division, where the SHAP algorithm is applied to unseen data in order to determine the shape values, and these values are further used to understand the trained model in detail. The SHAP algorithm uses the trained model and its tokenizer to determine SHAP values from unseen data.

\subsection{Code Availability}
The related codes of this paper is available  in https://github.com/Taharat22/Math\_entity\_Relationship.
The main experimental code, for example, the model design, result generation, and SHAP experiment, will be found in the notebook file named "English Explainable shap code for MER.ipynb." There are some other notebook files for drawing the graphs and other graphics. The final processed data can be found in "English ME Relationship WO tokens" file. Anyone using the resources from this public repository can reuse the code.

\section{Results}
\label{result}
\begin{figure*}
    \centering
    \includegraphics[width=\linewidth]{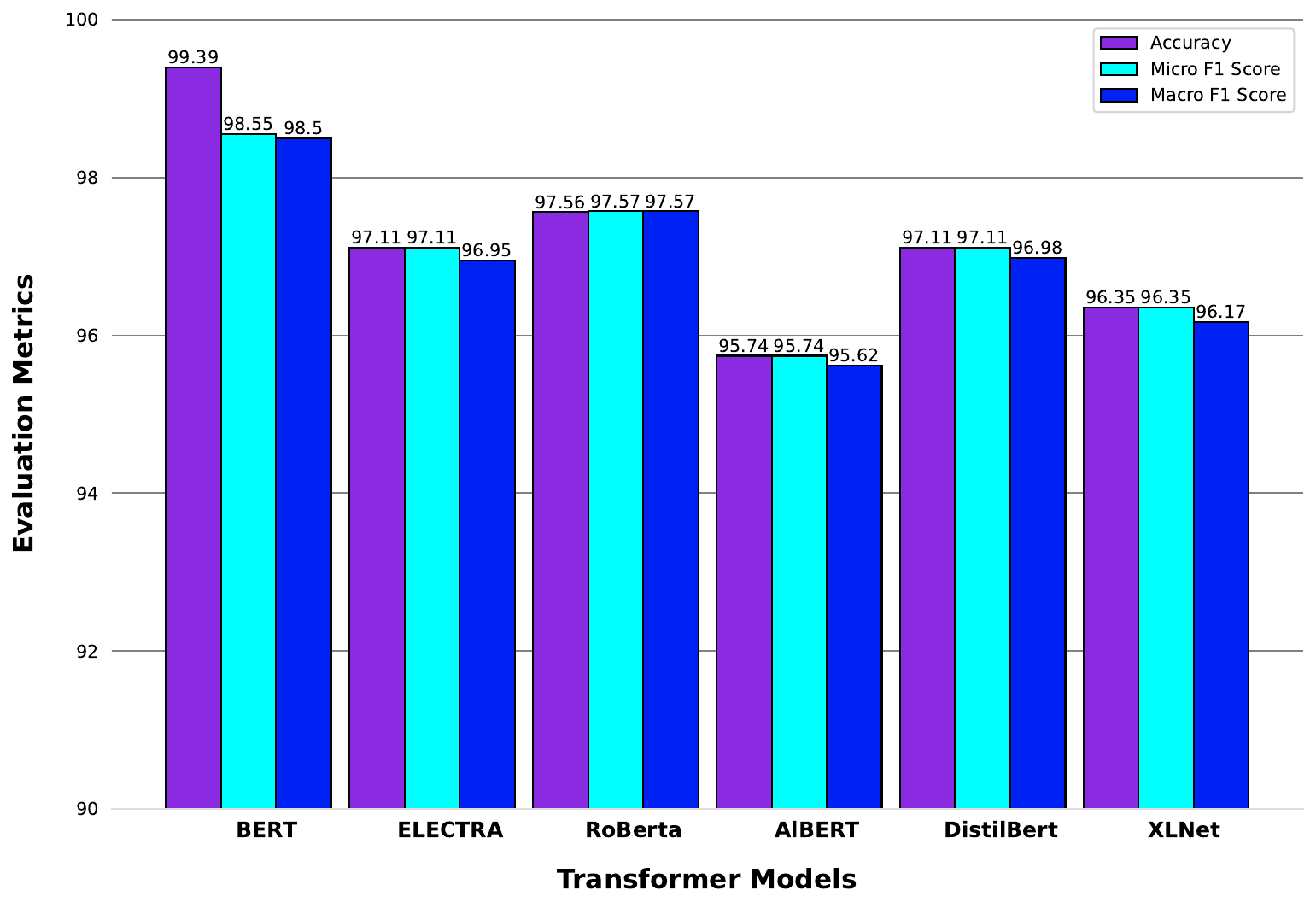}
    \caption{Comparative performance of transformer-based models (BERT, ELECTRA, RoBERTa, ALBERT, DistilBERT, and XLNet) in terms of accuracy, micro-average F1 score, and macro-average F1 score, demonstrating the superior performance of BERT.}
    \label{fig:comparisn}
\end{figure*}

Different transformer-based architectures like Electra, RoBERTa, AlBERT, DistillBERT, and XlNet are applied in the preprocessed dataset to select a model for this research. During training, the corresponding tokenizer is utilized for each transformer-based model.
Three different evaluation metrics, Accuracy Macro, and Micro average F1 scores, are calculated for the comparison. A bar chart is illustrated in Figure \ref{fig:comparisn}, including the metrics values for all the mentioned models. Visibly, it is clear that BERT provided the scores in the graph. The Accuracy Macro and Micro average F1 scores for BERT are respectively 99.39\%, 99.36\%, and 99.27\%, which are better than all other mentioned models.

\begin{figure}
    \centering
    \includegraphics[width=0.8\linewidth]{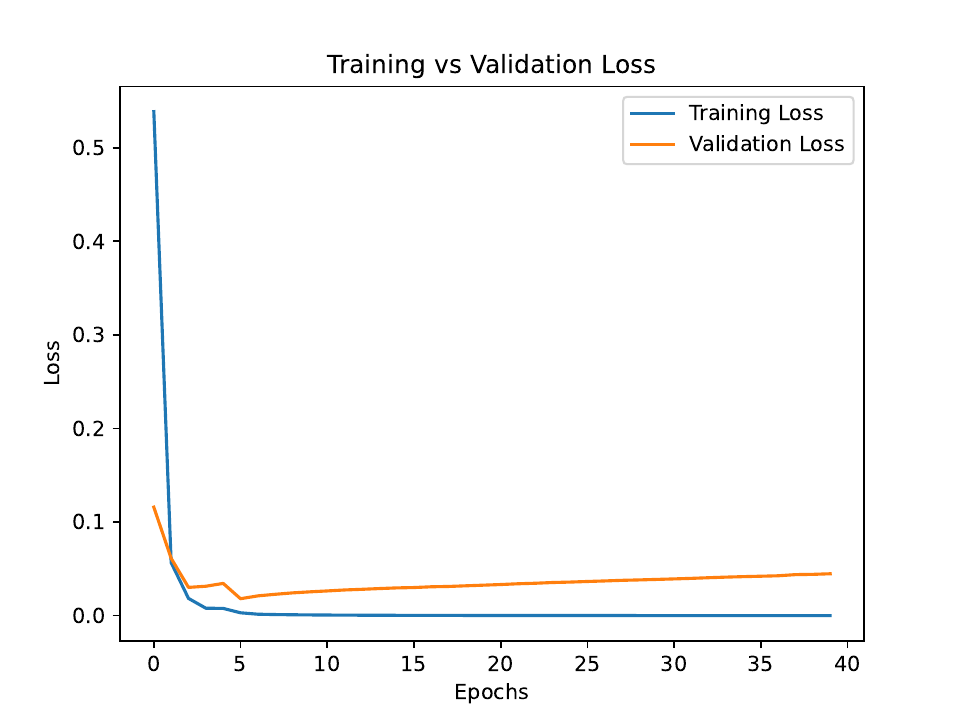}
    \caption{Training and validation loss curves across epochs for the proposed BERT-based model. The figure shows rapid convergence during the initial epochs, followed by stable training loss and a slightly increasing validation loss, indicating effective learning with minor overfitting at later stages.}
    \label{fig:lossover40}
\end{figure}

After comparing the transformer architecture, more outcomes are generated to justify BERT as the highest performer. Figure \ref{fig:lossover40} indicates the training and validation loss curves over 40 epochs for the BERT model.
The training and validation loss curves show a steady and consistent decrease across epochs, indicating effective model learning and stable optimization. The close alignment between training loss and validation loss suggests that the model achieves a good bias–variance trade-off, learning meaningful patterns from the data without overfitting. The absence of a widening gap between the two curves implies low variance, while the continuous reduction in loss indicates that the model is not underfitting (i.e., low bias). Overall, this behavior reflects a well-regularized model that generalizes effectively to unseen data, demonstrating balanced model capacity and reliable performance.

\begin{figure*}
    \centering
    \includegraphics[width=\linewidth]{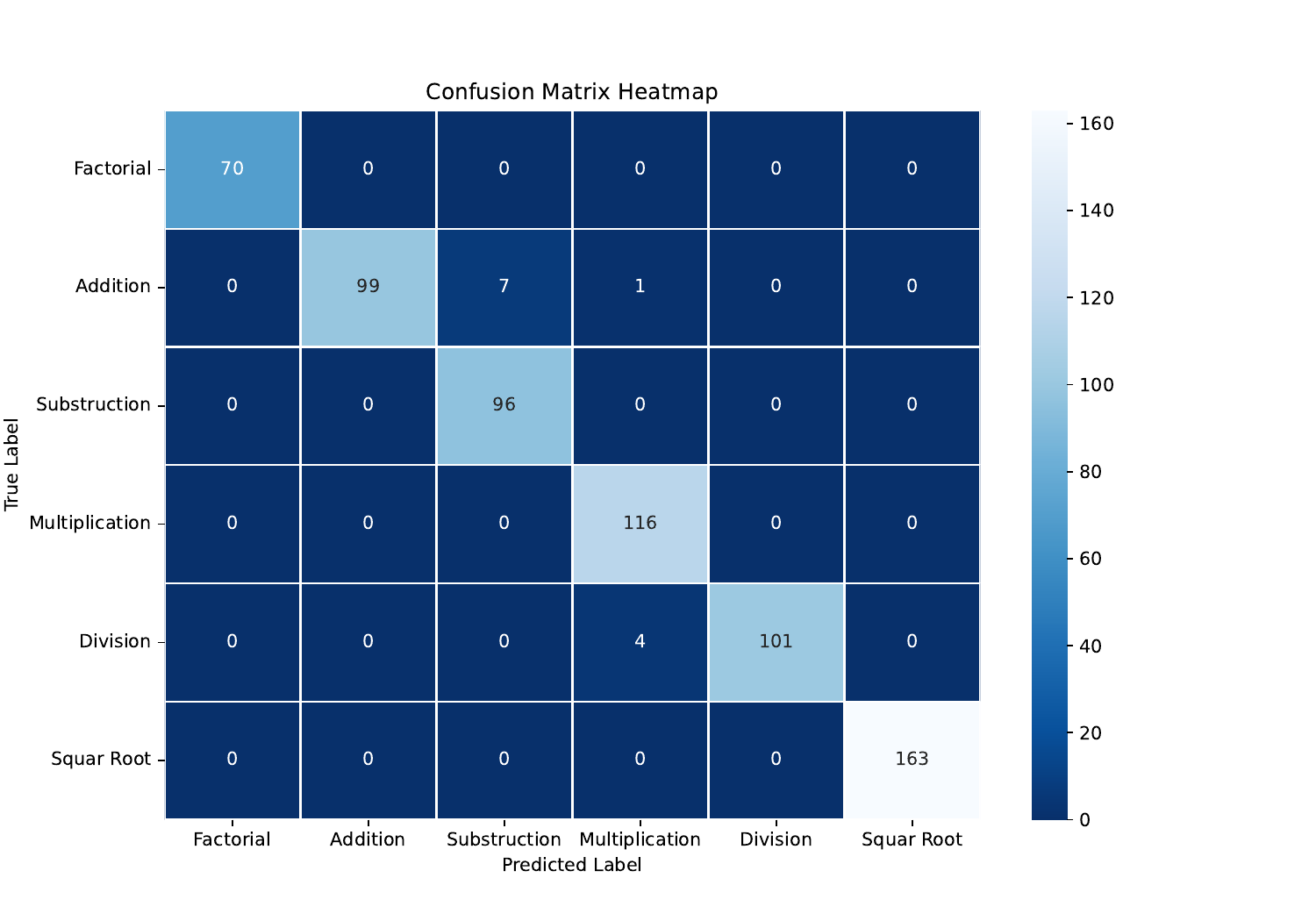}
    \caption{Confusion matrix heatmap for the BERT model on the test dataset, visualizing correct classifications and misclassifications across six mathematical relation classes.}
    \label{fig:conf}
\end{figure*}

A confusion matrix is plotted as a heatmap in Figure \ref{fig:conf} using a test dataset. Except for a few observations, in most cases, the model is capable of predicting the mathematical entity relationship precisely. For the relationship 'Addition,' the highest number of eight misclassifications occurred. In seven classes, 'Addition' is predicted as 'Subtraction,' and in one case, the model predicted 'Multiplication' instead of 'Addition.' Next, misclassification is observed in the prediction of the 'Division' math relationship. Four observations, which are supposed to be predicted as 'Division,' are misclassified as 'Multiplication.'

The obtained confusion matrix from Table \ref{tab:evaluation} is further utilized to determine different Evaluation metrics, for instance, Precision, Recall, Specificity, Error Rate, and F1 Score. For Precision, Recall, and F1 Score, two different approaches, Micro and Macro Average, are used. In Table \ref{tab:evaluation}, all values indicate that the model classifies the mathematical relationship correctly. On the other hand, the error rate is 0 or very low, as the model performed on 13 misclassifications among 657 observations. Therefore, the proposed model achieved outstanding performance in predicting the mathematical relationship from text.

\begin{table}[]
\centering
\caption{Different Evaluation Metrices for the BERT model}
\label{tab:evaluation}
\begin{tabular}{|l|c|c|c|c|c|c|c|c|c|}
\hline
\textbf{Class} & \textbf{\begin{tabular}[c]{@{}c@{}}TP\end{tabular}} & \textbf{\begin{tabular}[c]{@{}c@{}}TN\end{tabular}} & \textbf{\begin{tabular}[c]{@{}c@{}}FP\end{tabular}} & \textbf{\begin{tabular}[c]{@{}c@{}}FN\end{tabular}} & \textbf{Precision} & \textbf{Recall} & \textbf{Specificity} & \textbf{\begin{tabular}[c]{@{}c@{}}Error\\ Rate\end{tabular}} & \textbf{\begin{tabular}[c]{@{}c@{}}F1 \\ Score\end{tabular}} \\ \hline
\textbf{Factorial} & 70 & 466 & 0 & 0 & 100 & 100 & 100 & 0 & 100 \\ \hline
\textbf{Addition} & 99 & 459 & 8 & 1 & 93 & 99 & 98 & 1.8 & 96 \\ \hline
\textbf{Subtraction} & 96 & 467 & 0 & 2 & 100 & 98 & 100 & 0.4 & 99 \\ \hline
\textbf{Multiplication} & 116 & 449 & 4 & 0 & 97 & 100 & 99 & 0.8 & 98 \\ \hline
\textbf{Division} & 101 & 454 & 0 & 4 & 100 & 96 & 100 & 0.8 & 98 \\ \hline
\textbf{Square Root} & 163 & 392 & 0 & 0 & 100 & 100 & 100 & 0 & 100 \\ \hline
\textbf{\begin{tabular}[c]{@{}l@{}}Macro \\ Average\end{tabular}} & - & - & - & - & 98.5 & 98.5 & 99.5 & 0.63 & 98.5 \\ \hline
\textbf{\begin{tabular}[c]{@{}l@{}}Micro \\ Average\end{tabular}} & - & - & - & - & 98.3 & 98.9 & - & - & 98.6 \\ \hline
\end{tabular}
\end{table}

\begin{figure*}
    \centering
    \includegraphics[width=1.1\linewidth]{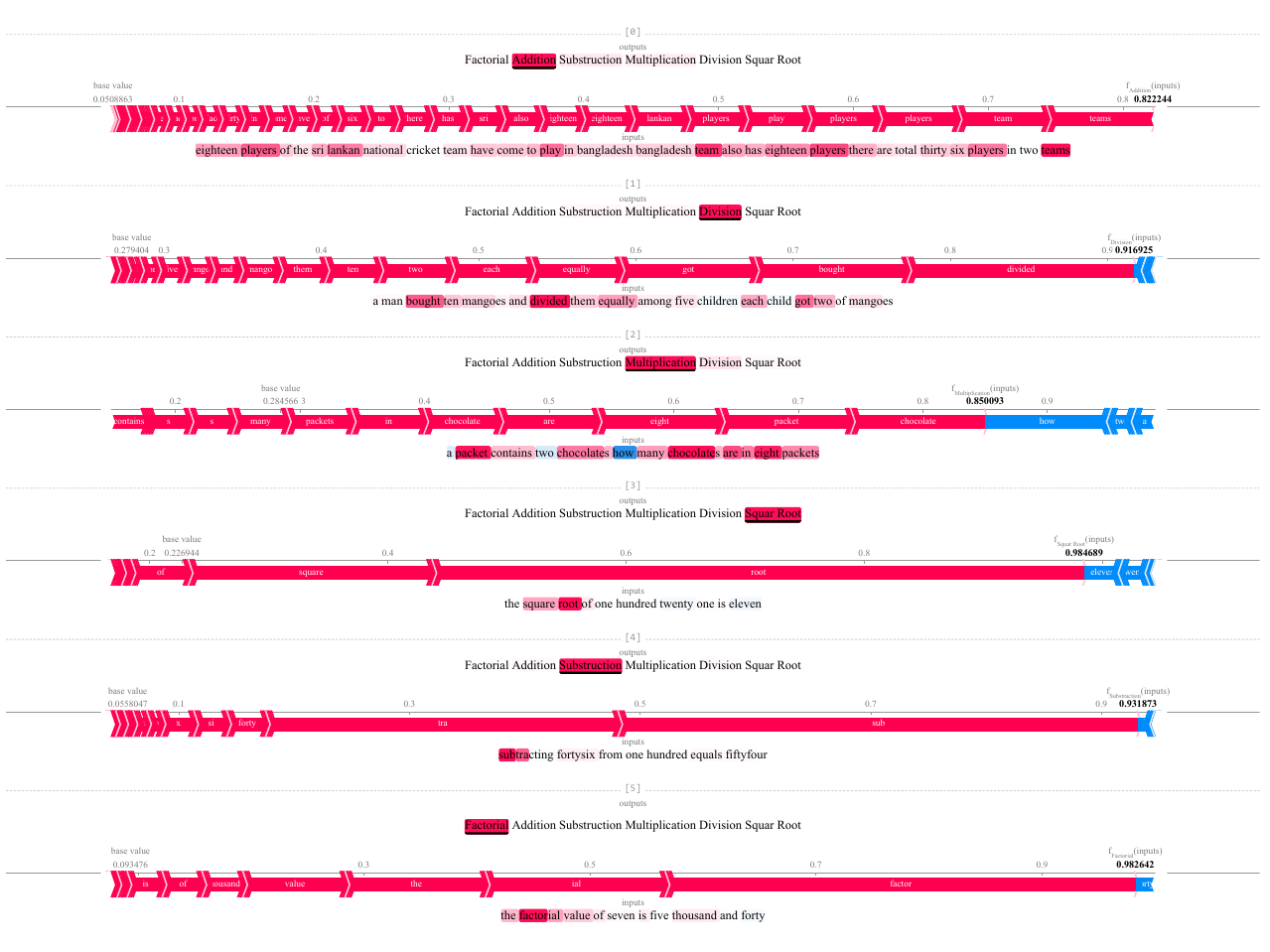}
    \caption{SHAP text plots illustrating word-level feature contributions for predicting distinct mathematical relation classes, where red highlights indicate positive contributions and blue highlights indicate negative contributions.}
    \label{fig:shaptext}1
\end{figure*}

A list of math texts containing observations from each class is passed on to the SHAP Explainer. The SHAP explainer, by maneuvering the trained model and its tokenizer, provides SHAP values. These SHAP values are then used to understand the important features of the model's prediction. Figure \ref {fig:shaptext} is a plot developed from these SHAP values and shows the vital word features for predicting each Math Relation. The figure shows which word how much contributed to the model's prediction for each class. The SHAP explainer not only perfectly determines the output classes but also explains the important features of the prediction.

Figure \ref{fig:shaptext} uses SHAP values to indicate each word has an influence on the model's prediction for a specific mathematical relationship (like Division, Multiplication, or Square Root). Each row represents a different word problem sentence, showing the tokens that the model has processed.
Some words are highlighted in red or blue. Red shading indicates a positive contribution toward a particular output relationship, while blue indicates a negative contribution to that relationship. Blue words are tokens that decrease the probability of the model predicting the target operation. For example, in a row where the model predicts "Division" as the operation, blue words have a negative SHAP value, meaning they contribute less or even reduce the confidence for "Division.
In the first observation where the output is 'Addition,' the prediction is calculated with the significant contribution of 'eighteen,' 'players,' 'team,' and 'thirty-six.'  Similarly, in the second observation, the 'Division' class is predicted utilizing the tokens obtained from the words 'divided,' 'equal,' 'bought,' and 'each.' 
Successive observations of the importance of a word's contributions are similarly measured in Figure \ref{fig:shaptext}. 

\begin{figure*}[ht!]
    \centering
    \includegraphics[width=\linewidth]{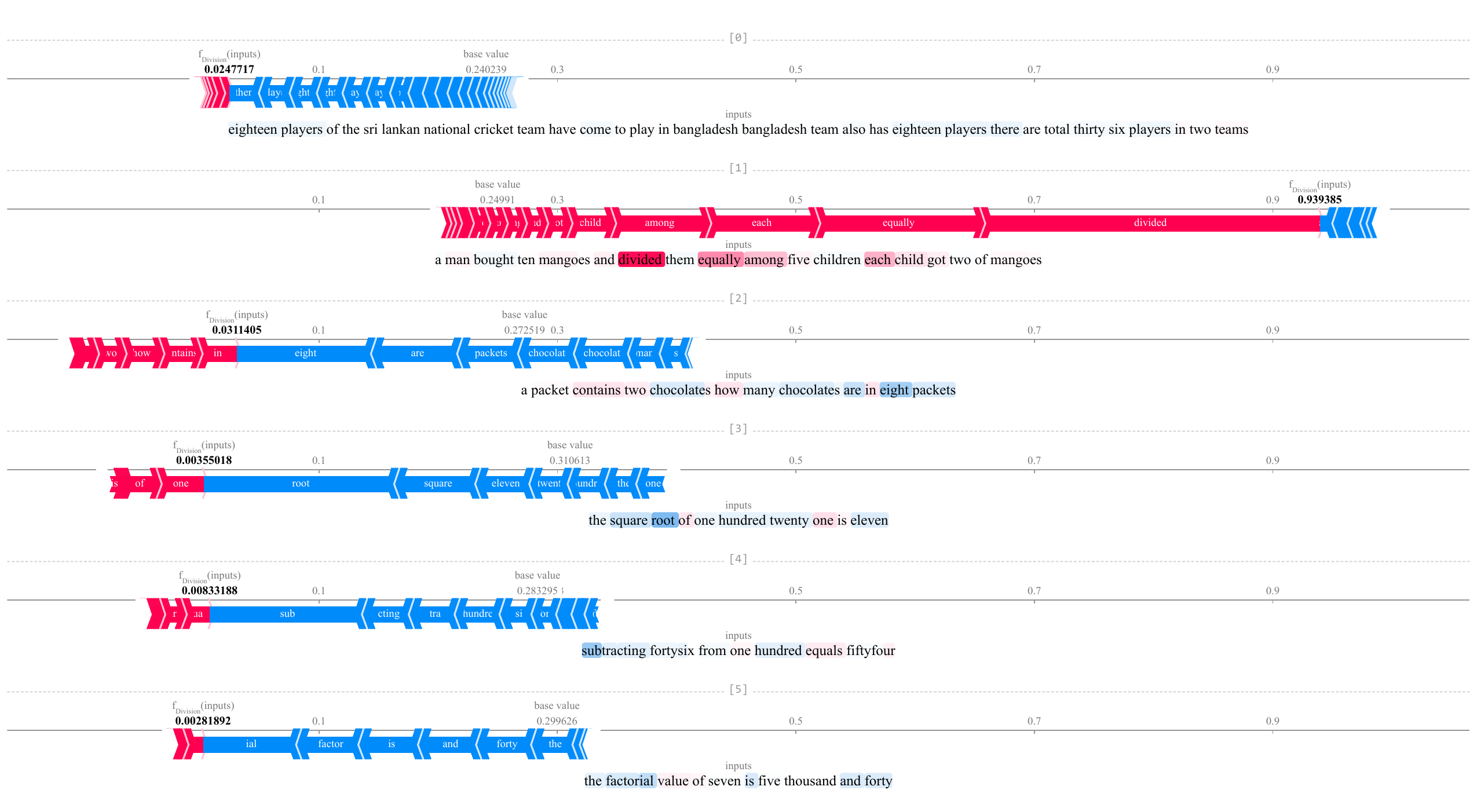}
    \caption{SHAP-based word contribution analysis for the “Division” class across multiple observations, showing how feature importance influences prediction confidence for both relevant and irrelevant samples.}
    \label{fig:shapclasswise}
\end{figure*}

A base value is visible on the left side of each row, and a final SHAP value($f_x$ where $x$ indicates the relationship name) is on the right side of the row. 
The base value in a SHAP plot represents the model's initial prediction before considering the impact of any specific features. It is a baseline that captures the average model output across all instances in the dataset. As individual feature contributions (SHAP values) are added or subtracted from this base value, they adjust the prediction toward the final model output for a specific instance. Finally, the SHAP value $f_x$ indicates the confidence of a class to be the predicted output for that observation. For example, in the first row, the base value is 0.0508863, which means the initial probability of 'Addition' being the outcome of the first row is 0.0508863 without considering any feature aspects. After considering different features and their SHAP values, the final probability of 'Addition' being the outcome is $f_{Addition}\ =$ 0.822244 or 82\%. From the next rows, it is visible that a smaller base value determines higher confidence for the predicted class. In Figure \ref{fig:shapclasswise}, the confidence of the six-row outcome is higher than 80\%. From first to last, the confidence of predicted relationships Addition, Division, Multiplication, Square Root, Subtraction, and Factorial are 82\%, 92\%, 85\%, 98\%, 93\%, 98\%. These values are higher to assist the model in predicting the correct relation.
While the base value is the initial prediction of the model, the shape value leads to the final prediction.

In previous Figure \ref{fig:shapclasswise}, the contribution of the feature from different observations for the respective predicted class is illustrated. Now each observation feature contribution, only one single class will be discussed here, and the 'Division' class is randomly chosen for this discussion. Figure \ref{shap classwisedivision.pdf} depicted the contribution of features from each row to determine the confidence of the 'Division' class to be the predicted outcome. Only for the second row is the correct prediction in the 'Division' class, and the remaining observations are not supposed to be related to this class. Therefore, Then, randomly one class is selected, and the contribution of each observation's features is calculated to determine the confidence of the 'Division' class in Figure \ref{fig:shapclasswise}. For six distinct text observations, the confidences of 'Division' are 2\%, 94\%, 3\%, 4\%, 8\%, and .2\%. Only for the second observation is the confidence value higher, as the true label of this text is 'Division.'

\begin{figure*}[ht!]
    \centering
    \includegraphics[width=\linewidth, height=.8\textheight]{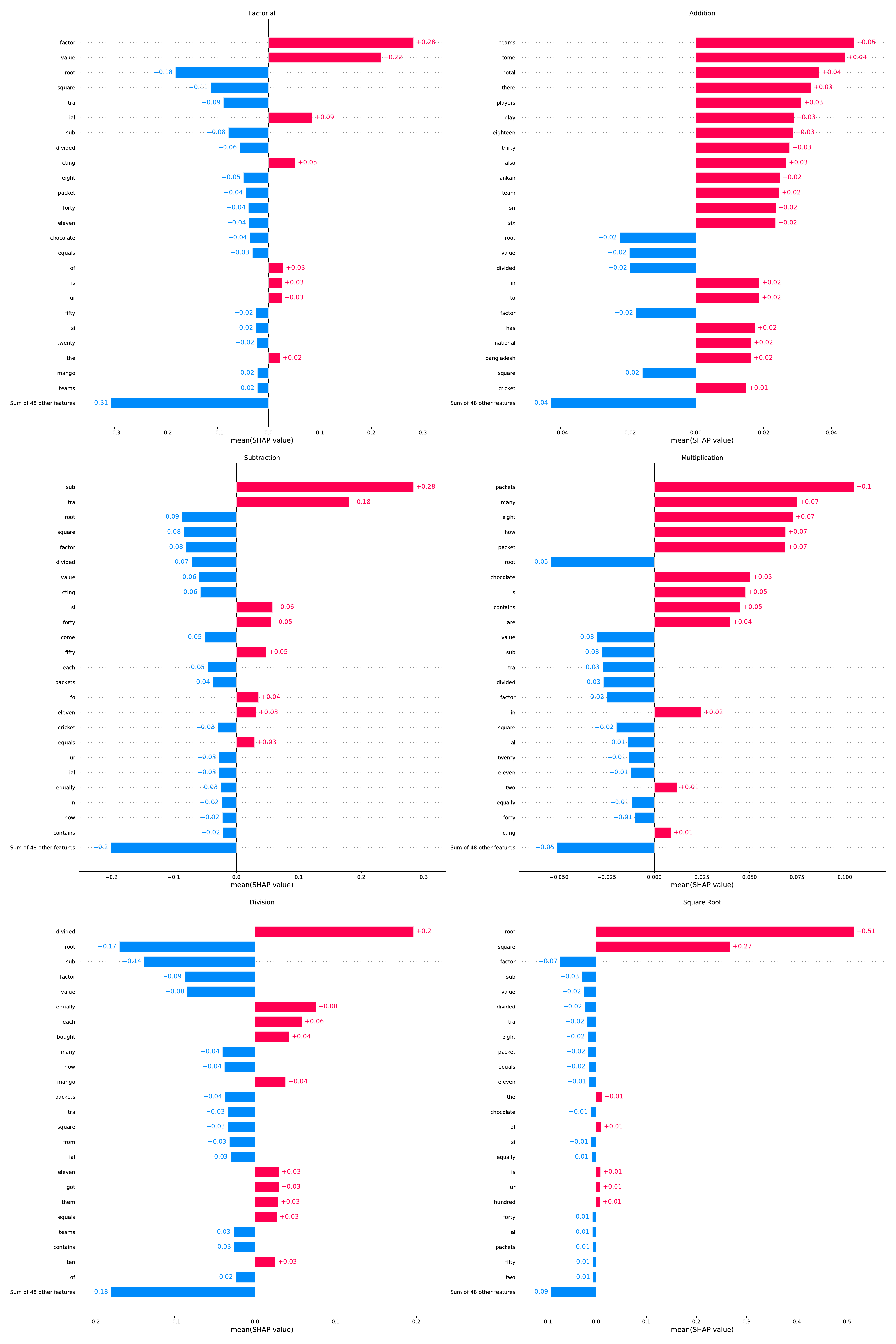}
    \caption{SHAP bar plots presenting mean feature importance values for each mathematical relation class, revealing operation-specific keywords and distributed contextual contributions that drive model predictions.}
    \label{fig:shapbar}
\end{figure*}

For more clarification, the Figure \ref{fig:shapbar} is added. In this figure, SHAP bar plots are illustrated for all classes. 
Across all mathematical operation classes, the model’s predictions are driven primarily by \textbf{operation-specific keywords rather than isolated numeric values}. Words that directly signal the underlying mathematical intent—such as divide, root, square, factor, sub, and equals—consistently dominate the contribution space. This indicates that the model has learned to associate semantic cues of operations more strongly than surface-level numbers, suggesting a solid grasp of mathematical language patterns rather than memorization of particular values.
For \textbf{Division and Square Root}, the model shows especially strong confidence. Square Root stands out with a very high mean SHAP contribution from keywords like root and square, implying that these terms are highly distinctive and rarely confused with other operations. Similarly, Division is heavily influenced by words such as divided, equally, and each, reflecting real-world distribution contexts (e.g., packets, mangoes, chocolates) that clearly anchor the mathematical meaning. This suggests that the dataset provides strong contextual signals for these operations, allowing the model to separate them cleanly from others.
In contrast, \textbf{Addition, Subtraction, and Multiplication} rely on a broader mix of contextual and relational words. Terms indicating grouping, totals, comparison, or change—such as total, many, contains, how, and equals—play a collective role rather than a single dominant feature. This implies that these operations are linguistically more ambiguous and often require multiple cues to disambiguate, which the model addresses by aggregating evidence across several features rather than relying on one decisive keyword.
Another important observation is the \textbf{substantial contribution from the “sum of other features”} across all classes. This indicates that while top keywords are influential, the model also depends on a wide tail of secondary words and contextual signals. Such behavior is desirable, as it reflects a distributed understanding of problem statements rather than over-dependence on a few trigger words, improving robustness and generalization.
Overall, the SHAP analysis demonstrates that the model has learned \textbf{operation-aware semantic representations}, effectively mapping linguistic structure to mathematical intent. Strong separability for operations like Square Root and Division, combined with more context-driven reasoning for Addition and Subtraction, aligns well with how humans interpret math word problems. This provides confidence that the model’s predictions are not only accurate but also \textbf{interpretable and linguistically grounded}.

\section{Discussion}
This paper represents an approach to mathematical problem-solving, considering a math statement as an entity relationship problem. The operands are the entities, and the operation performed between them is the entity relationship. This work was conducted by utilizing explainable deep learning to understand and explain the relation between two math entities, specifically operands.
By using transformer-based methods to extract relationships between mathematical entities, the way for exciting practical applications has been paved. This includes creating smart educational tools that offer precise answers to math questions, developing systems that can automatically check complex proofs in mathematical research, and building detailed knowledge maps to help people better navigate and understand complex math topics. This research has the potential to greatly improve the speed and accuracy of working with mathematical information, benefiting both education and professional work.

The data for this research is collected by combining two datasets, and the sample observations are detailed in Table \ref {tab:dataset}. The combined dataset consists of three columns: Text, Entity 1, Entity 2, and the Relationship. Six relationships are determined, namely Factorial, Addition, Subtraction, Multiplication, Division, and Square Root. These are the most basic and helpful math operations.
Next, the text is preprocessed, and Figure \ref{fig:data preprocess} represents the entire data preprocessing with an example. Here, the text "A man bought ten mangoes, and divided them equally among five children. Each child got two of the mangoes."  is step-by-step preprocessed by applying character removal, stop word removal, lemmatization, and stemming.

In this research, all popular transformer-based models BERT, Electra, RoBERTa, AlBERT, DistillBERT, and XlNet are measured and compared. For this comparison, three significant evaluation metrics are utilized: accuracy, micro, and macro F1 scores. Figure \ref{fig:comparisn} represents this comparison, and All the models provided remarkable scores which are larger than 95\%. The outstanding performance of the BERT model is visible from this bar graph. Therefore, BERT was chosen to analyze the results further. The confusion matrix of the BERT classifier is depicted as a Heatmap in Figure \ref{fig:conf} where the correct classifications are visibly higher than the misclassifications. Utilizing this confusion matrix, more scores are measured. In table \ref{tab:evaluation}, the value of Precision, Recall, Specificity, Error Rate, and F1 scores for six math relations are determined. The Error Rate is visibly lower in values, even very close to zero for each class, and the remaining score is significantly higher, which is almost 100, eventually 100 for some classes.

Deep learning models, with their complex architecture, often function as black boxes, making it hard to know how features are calculated to determine the outputs. While the proposed model has achieved remarkable performance, it's equally important to understand how and why this performance is achieved. This is where the XAI algorithm SHAP comes in. By bringing the black box to light, SHAP helps us understand which features are important for the prediction and which lead the model to perform errors, empowering us with a deeper understanding of our models. 
Figure \ref{fig:shaptext} is a text plot that highlights the most significant features of each math relationship class. The red highlighted words are the positive, and the blue highlighted words are the negative contribution for calculating the confidence of the class. 
Then, randomly selected one class and calculated the contribution of each observation's features to determine the confidence of the 'Division' class in Figure \ref{fig:shapclasswise}.
Furthermore, Figure \ref{fig:shapbar} provides SHAP bar plots for each math relationship, which indicate the positive(red) and negative(blue) contributions of different words in calculating the probability of that class.
A more detailed description is already provided in Section \ref{result} along with each Figure.
\subsection{Limitation}
Though this work intends to present a solution by utilizing modern deep learning techniques, it has some limitations yet to be addressed. Following limitations. In the future, we want to address the following limitations: 
\begin{itemize}
    \item Dataset: The model shows higher accuracy on the downstream task. The texts have different variations of real-world math problems. However, the text used here is straightforward and small in length.
The dataset is limited to only six basic equations, and for now, the work is not focused on complex math problems. In the future, this work intends to address more complex mathematical problems using longer sequences.
\item Model's Efficiency: The proposed model approach is computationally heavy as we are using the BERT and SHAP at the same time. In the future, we want to optimize the model.
\end{itemize}

\section{Conclusion and Future Work}
This research work highlights the role of mathematics in the field of education and research. Moreover, it emphasizes its importance in advancing science and engineering. It demonstrates the increasing demand for modern tools that can efficiently tackle math problems, reflecting the fast pace of technological growth. By approaching math problems as relation extraction tasks, automated systems can be able to interpret and analyze complex mathematical expressions. This can help to develop advanced applications like automated theorem proving, answering math questions, and building knowledge graphs for mathematical content. Our use of transformer-based models, especially BERT, has been highly successful, reaching an impressive accuracy of 99.39\%, instilling confidence in the advancements in AI and math problem-solving. Additionally, to ensure our model’s transparency and explainability,  XAI methods are incorporated, precisely SHAP, to clarify the model’s performance, essential features, and any errors.

Future work will expand this research to other areas of mathematics, such as algebra, geometry, and calculus, to better understand how well the model works across different fields. Further research should aim to connect these automated systems with advanced tools, like automated theorem proving and creating detailed knowledge graphs for math content. Collaboration with experts in fields like computer science, linguistics, and education could help build more robust and flexible automated solutions.

\section{Ethics declarations}

\subsection{Competing Interests:} The authors declare no competing interests.
\subsection{Declarations:} Not applicable
\subsection{Ethical Approval:} Not applicable. This study does not involveData Availability human or animal subjects, and therefore, no ethical approval was required.
\subsection{Funding: } No funding was received for conducting this study.
\section{Data and Code Availability }
The dataset will be found in-\\
https://www.sciencedirect.com/science/article/pii/S2352340924003767 \\
https://www.sciencedirect.com/science/article/pii/S2352340924007091

\bibliography{cas-refs.bib}

\end{document}